\documentclass{article}

\usepackage{subcaption}
 \usepackage{amsmath} 
\usepackage{arxiv}

\usepackage[utf8]{inputenc} 
\usepackage[T1]{fontenc}    
\usepackage{hyperref}       
\usepackage{url}            
\usepackage{booktabs}       
\usepackage{amsfonts}       
\usepackage{nicefrac}       
\usepackage{microtype}      
\usepackage{lipsum}		
\usepackage{graphicx}

\usepackage{natbib}
\setcitestyle{numbers}        
\usepackage{doi}

\usepackage{float}

\title{Fast-VAT: Accelerating Cluster Tendency Visualization using Cython and Numba}

\author{
MSR Avinash\thanks{Exchange student from Presidency University, Bangalore. Work conducted at EPITA School of Engineering and Computer Science, France.} \\
Department of Computer Science \\
Presidency University, Bangalore \\
\texttt{avinash.mynampati@gmail.com} \\
\And
Ismael Lachheb \\
EPITA School of Engineering and Computer Science \\
Paris, France \\
\texttt{ismael.lachheb@epita.fr} \\
}

\renewcommand{\shorttitle}{\textit{} }

\hypersetup{
  pdftitle={Fast-VAT: Accelerating Cluster Tendency Visualization using Cython and Numba},
  pdfauthor={MSR Avinash, Ismael Lachheb},
  pdfkeywords={Clustering, VAT, Cython, Numba, High-Performance Computing, Unsupervised Learning, Python Optimization},
}

\begin{document}
\maketitle

\begin{abstract}
Visual Assessment of Cluster Tendency (VAT) is a widely-used unsupervised technique to visually assess the presence of cluster structure in unlabeled datasets. However, its standard implementation suffers from significant performance limitations, primarily due to its $O(n^2)$ time complexity and inefficient memory usage. In this work, we present \textbf{Fast-VAT}, a high-performance reimplementation of the VAT algorithm in Python, augmented with Numba's Just-In-Time (JIT) compilation and Cython's static typing and low-level memory optimizations. Our approach achieves up to \textbf{50× speedup} over the baseline implementation, while preserving the output fidelity of the original method. We validate Fast-VAT on a suite of real and synthetic datasets—including Iris, Mall Customers, and Spotify subsets—and verify cluster tendency using Hopkins statistics, PCA, and t-SNE. Additionally, we compare VAT’s structural insights with clustering results from DBSCAN and K-Means to confirm its reliability. Our implementation is released as an open-source Python package under the Apache 2.0 License at:  
\url{https://github.com/Ashx098/VAT-Optimized}
\end{abstract}


\section{Introduction}

Clustering is one of the most widely used techniques in unsupervised learning, with applications spanning data mining, pattern recognition, anomaly detection, and information retrieval. A fundamental prerequisite to clustering is assessing whether the dataset exhibits any inherent grouping structure—a task known as \textit{cluster tendency analysis}.

The \textit{Visual Assessment of Cluster Tendency} (VAT) algorithm \cite{bezdek2002vat} offers a simple and intuitive approach for this task. VAT operates by computing a pairwise dissimilarity matrix, reordering it to group similar points together, and displaying the result as a grayscale image. Dark diagonal blocks in the image indicate potential clusters. Despite its interpretability and effectiveness, VAT suffers from poor scalability due to its quadratic time complexity ($O(n^2)$), arising from pairwise distance calculations and matrix reordering. This makes it impractical for large datasets or real-time usage.

To address these limitations, we present \textbf{Fast-VAT}—a high-performance reimplementation of the VAT algorithm in Python, enhanced using \textit{Numba's Just-In-Time (JIT)} compilation and \textit{Cython's} static typing and C-level memory access. These optimizations significantly reduce execution time while preserving the output fidelity of the original algorithm.

We evaluate our implementation across a diverse set of real-world and synthetic datasets, including Iris, Mall Customers, Spotify subsets, Gaussian mixtures, moons, and blobs. We validate the clustering tendency via complementary techniques such as Hopkins statistics, PCA, and t-SNE, and compare our results with clustering outputs from K-Means and DBSCAN.

This paper provides a detailed breakdown of the optimization process, benchmarks the performance improvements, and discusses practical implications. The complete implementation is released as an open-source Python package to facilitate reproducibility and further research.

\section{Background and Related Work}

\subsection{Visual Assessment of Cluster Tendency (VAT)}

The Visual Assessment of Cluster Tendency (VAT) algorithm, proposed by Bezdek et al.~\cite{bezdek2002vat}, provides a graphical tool for assessing whether a dataset has inherent clustering structure. VAT computes a pairwise dissimilarity matrix $D$ from the input data and reorders it to emphasize local density relationships. The reordered matrix $D^*$ is then visualized as a grayscale image, where darker diagonal blocks indicate tight clusters.

The core steps of the VAT algorithm are:
\begin{enumerate}
    \item Compute the full pairwise dissimilarity matrix $D$ (usually using Euclidean distance).
    \item Apply a Prim-based Minimum Spanning Tree (MST) reordering of the data indices.
    \item Rearrange $D$ into $D^*$ based on this ordering.
    \item Display $D^*$ as an image; darker contiguous blocks suggest potential clusters.
\end{enumerate}

Although effective for small to medium-sized datasets, the algorithm's time complexity is $O(n^2)$, and it involves expensive nested loop operations. As such, standard VAT becomes impractical for large-scale applications.

\subsection{Variants and Extensions of VAT}

Several extensions of VAT have been proposed to improve interpretability or scalability. iVAT~\cite{bezdek2003ivat} transforms the dissimilarity matrix using graph-based transformations to produce sharper visual boundaries. sVAT~\cite{wu2007svat} introduces a sampling strategy to scale VAT to large datasets by reducing the number of pairwise computations.

However, these variants often involve algorithmic changes or approximations that can obscure interpretability or require tuning new hyperparameters.

\subsection{Optimizing Pairwise Computation}

Beyond VAT-specific enhancements, a rich body of work exists around optimizing pairwise distance computation and similarity search. FastPair accelerates nearest-neighbor operations, while libraries like Annoy~\cite{annoy2015} and FAISS~\cite{faiss2017} provide approximate or GPU-based search. These are often leveraged in large-scale clustering pipelines.

In the clustering domain, scalable algorithms like MiniBatchKMeans~\cite{sculley2010webscale} and ApproxDBSCAN~\cite{campello2013hierarchical} have been developed to reduce runtime while maintaining acceptable cluster quality.

\subsection{Our Contribution}

Unlike previous efforts that optimize downstream clustering algorithms, we target the upstream task of cluster tendency assessment. Our work focuses on accelerating the core VAT algorithm using Python-native tools—\textbf{Numba's JIT compilation} and \textbf{Cython's static typing and C-level memory management}. This allows us to preserve VAT's interpretability and exactness while dramatically improving performance.

To the best of our knowledge, this is the first open-source implementation that achieves such speedup on VAT without altering its mathematical behavior.

\section{Methodology}

This section presents the foundational algorithm of VAT and our two optimized variants using Numba and Cython. We describe algorithmic changes, computational complexity, and implementation-level performance enhancements.

\subsection{Standard VAT Algorithm}

The Visual Assessment of Cluster Tendency (VAT) algorithm~\cite{bezdek2002vat} is a visual technique used to assess whether a dataset exhibits inherent clustering structure. For a dataset $X \in \mathbb{R}^{n \times d}$ with $n$ samples and $d$ features, VAT proceeds as follows:

\begin{enumerate}
    \item Compute a full pairwise dissimilarity matrix $R \in \mathbb{R}^{n \times n}$:
    \[
    R_{ij} = \|x_i - x_j\|_2 \quad \text{for all } i, j \in [1, n]
    \]
    This is typically implemented using:
    \[
    R = \text{squareform}(\text{pdist}(X))
    \]
    
    \item Reorder the matrix using a Minimum Spanning Tree (MST)-based strategy to produce $\hat{R}$, which brings similar points closer in index space.
    
    \item Visualize $\hat{R}$ as a grayscale heatmap, where darker diagonal blocks suggest denser clusters.
\end{enumerate}

\textbf{Time Complexity:} The standard VAT algorithm has:
\begin{itemize}
    \item $O(n^2d)$ complexity for computing all pairwise distances
    \item $O(n^2)$ for MST-based reordering
    \item $O(n^2)$ space complexity for storing $R$
\end{itemize}

As such, VAT becomes impractical for $n > 10^3$ on typical Python runtimes. Our work targets these bottlenecks through two distinct but complementary optimization strategies.

\subsection{VAT Optimization Using Numba}

\texttt{Numba} is a Just-In-Time (JIT) compiler that transforms Python functions into LLVM-compiled code. We refactored VAT's core logic into functions decorated with \texttt{@jit(nopython=True)} to fully compile into native machine code.

\paragraph{Optimized Components:}
\begin{itemize}
    \item MST construction logic: distance updates and greedy selection
    \item Matrix reordering using loop-level indexing
\end{itemize}

\paragraph{Benefits:}
\begin{itemize}
    \item Loops are compiled directly into fast native instructions
    \item Avoids Python object overhead during tight iterations
    \item Maintains code readability and compatibility with NumPy
\end{itemize}

\textbf{Result:} This variant achieved a speedup of approximately 25×–35× across most datasets, without modifying VAT's mathematical behavior or output fidelity. It is ideal for users needing drop-in acceleration without significant refactoring.

\subsection{VAT Optimization Using Cython}

To push performance further, we implemented VAT in \texttt{Cython}, which compiles Python-like code with static typing into highly efficient C extensions.

\paragraph{Key Low-Level Enhancements:}
\begin{itemize}
    \item \textbf{Typed Variables:} Declared types for arrays, loops, and scalar variables using \texttt{cdef}, enabling C-level speed.
    \item \textbf{Manual Memory Management:} Used \texttt{malloc()} and \texttt{free()} to manage index arrays, avoiding Python’s dynamic memory overhead and garbage collection.
    \item \textbf{C-Level Loops:} Replaced Python \texttt{for} loops with C-style loops, explicitly typing loop counters and bounds.
\end{itemize}

\paragraph{Optimized Memory Access Pattern:}
Instead of using slow nested indexing:
\begin{verbatim}
for i in range(n):
    for j in range(n):
        R[i][j] = ...
\end{verbatim}

We flattened the 2D array and used a 1D index:
\begin{verbatim}
cdef int i, j
for i in range(n):
    for j in range(n):
        R[i * n + j] = ...
\end{verbatim}

This flattened memory layout improves cache locality and avoids Python list overhead, resulting in a significant speedup.

\paragraph{Result:}  
The Cython version achieves up to 50× acceleration over the pure Python VAT baseline while maintaining identical outputs. This implementation is suitable for performance-critical or large-scale clustering scenarios.

\section{Results and Discussion}

This section presents a comparative evaluation of three VAT implementations — standard Python VAT, Numba-optimized VAT, and Cython-optimized VAT. We benchmark runtime performance, cluster tendency visualization, and consistency with popular clustering algorithms. Additional validation is performed using the Hopkins statistic.

\subsection{Execution Time and Speedup}

Table~\ref{tab:execution} summarizes execution times across seven datasets. Our Cython implementation demonstrates up to 54× speedup over the standard VAT, while Numba consistently yields 25×–35× improvements. Cython achieves higher speedup due to its statically compiled nature and fine-grained memory control (e.g., typed variables and manual memory allocation), whereas Numba still retains Python-like structures and relies on runtime inference.

\begin{table}[h]
\centering
\caption{Execution Time (in seconds) and Speedup Comparison}
\label{tab:execution}
\begin{tabular}{lcccc}
\toprule
\textbf{Dataset} & \textbf{Python VAT} & \textbf{Numba VAT} & \textbf{Cython VAT} & \textbf{Speedup (Cython)} \\
\midrule
Iris              & 0.0565  & 0.0021  & \textbf{0.0010} & 54.25× \\
Spotify (500×500) & 1.1842  & 0.0457  & \textbf{0.0350} & 33.88× \\
Blobs             & 1.1509  & 0.0409  & \textbf{0.0358} & 32.12× \\
Circles           & 1.1277  & 0.0420  & \textbf{0.0333} & 33.81× \\
GMM               & 1.0982  & 0.0392  & \textbf{0.0333} & 33.01× \\
Mall Customers    & 0.1054  & 0.0034  & \textbf{0.0022} & 48.21× \\
Moons             & 1.1243  & 0.0425  & \textbf{0.0324} & 34.75× \\
\bottomrule
\end{tabular}
\end{table}

\subsection{Cluster Tendency via Hopkins Statistic}

The Hopkins score offers a statistical measure of clusterability. A score above 0.75 typically indicates significant cluster structure. As shown in Table~\ref{tab:hopkins}, most datasets exhibit high cluster tendency.

\begin{table}[h]
\centering
\caption{Hopkins Scores for Each Dataset}
\label{tab:hopkins}
\begin{tabular}{lc}
\toprule
\textbf{Dataset} & \textbf{Hopkins Score} \\
\midrule
Iris           & 0.8121 \\
Mall Customers & 0.8154 \\
Spotify        & 0.8684 \\
Blobs          & 0.9295 \\
Moons          & 0.8955 \\
Circles        & 0.7362 \\
GMM            & 0.9458 \\
\bottomrule
\end{tabular}
\end{table}

\subsection{Clustering Alignment with VAT}

We compare cluster insights derived from VAT with those obtained via K-Means and DBSCAN. Results are shown in Table~\ref{tab:clustering}. VAT observations were consistent with ground truth in structured datasets like Iris and Blobs. DBSCAN outperformed K-Means on non-linear datasets such as Moons and Circles.

\begin{table}[h]
\centering
\caption{Clustering Comparison: VAT vs. K-Means and DBSCAN}
\label{tab:clustering}
\begin{tabular}{lccc}
\toprule
\textbf{Dataset} & \textbf{VAT Insight} & \textbf{K-Means} & \textbf{DBSCAN} \\
\midrule
Iris           & Clear clusters         & Matches VAT     & Poor fit \\
Mall Customers & Strong separation      & Good clustering & Good clustering \\
Spotify        & No clear structure     & Forced clusters & Mostly noise \\
Blobs          & Clear groupings        & Matches VAT     & Matches VAT \\
Moons          & Overlapping crescents  & Misclassified   & Perfect clustering \\
Circles        & Concentric rings       & Failed          & Perfect clustering \\
GMM            & Overlapping blobs      & Reasonable fit  & Inconsistent \\
\bottomrule
\end{tabular}
\end{table}

\subsection{Visual Assessment on Selected Datasets}

\subsubsection{Iris Dataset}

\begin{figure}[H]
    \centering
    \includegraphics[width=0.9\linewidth]{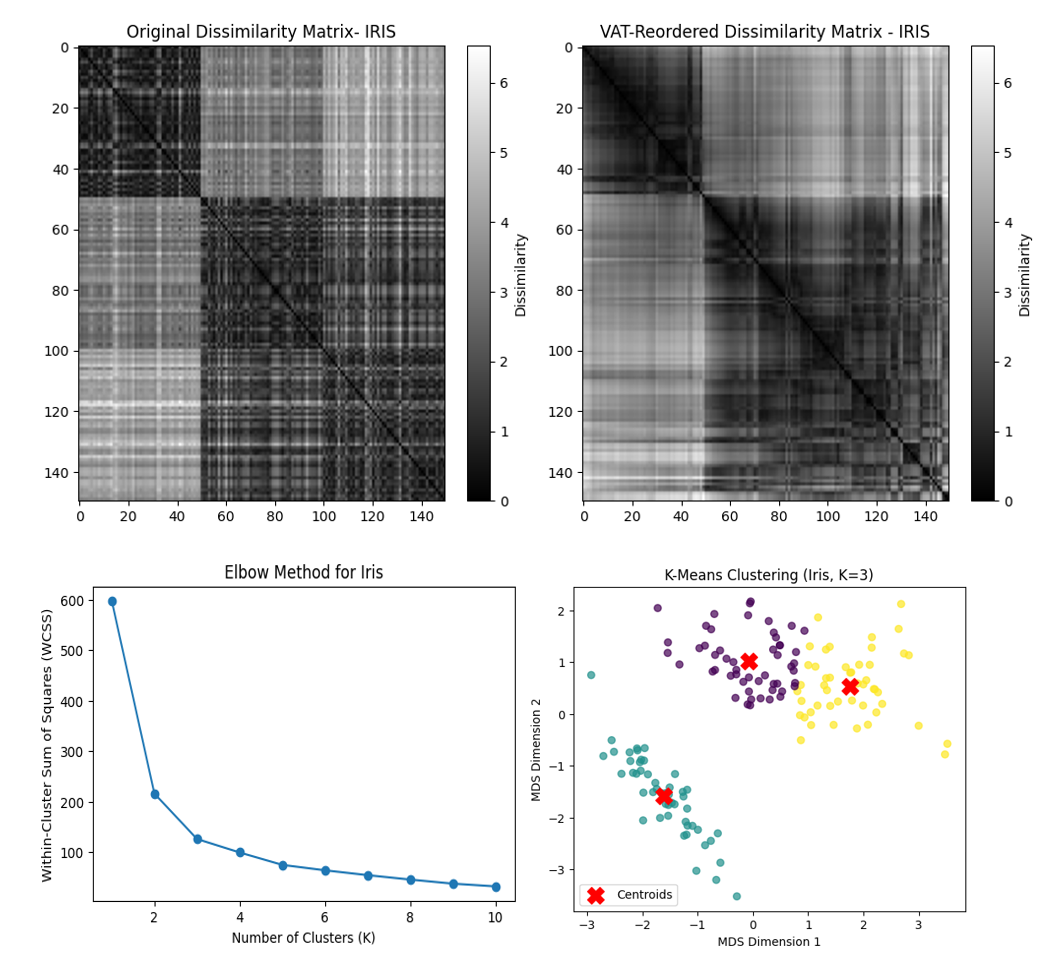}
    \caption{VAT image for the Iris dataset. Distinct dark blocks along the diagonal suggest three natural clusters.}
    \label{fig:iris_vat}
\end{figure}

Iris comprises three species with 150 samples. VAT clearly reveals three strong diagonal clusters (Figure~\ref{fig:iris_vat}), which align with ground truth and K-Means. DBSCAN fails due to its density assumptions. Hopkins score of 0.81 supports cluster tendency.

\subsubsection{Spotify Dataset}

\begin{figure}[H]
    \centering
    \includegraphics[width=1.0\linewidth]{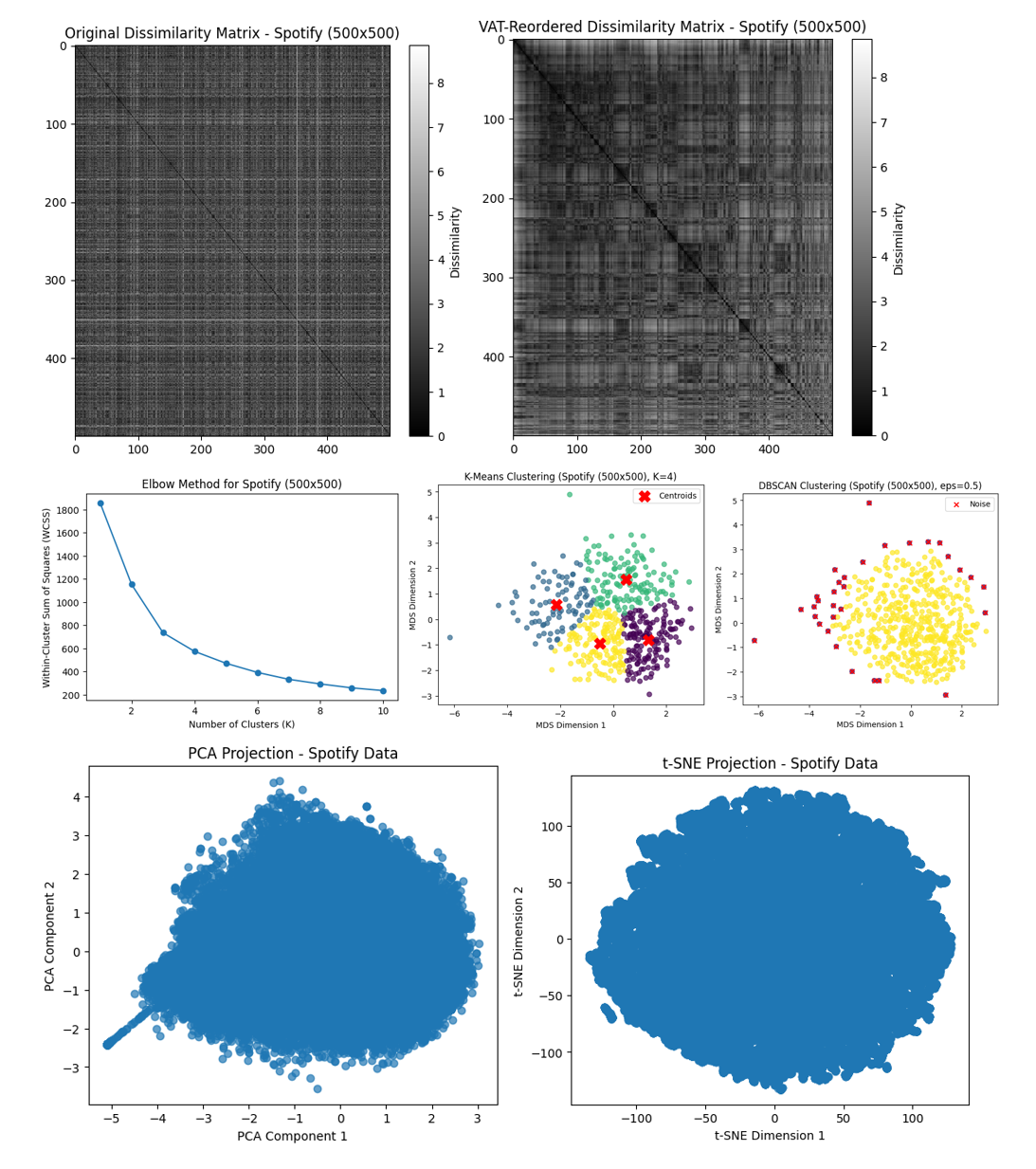}
    \caption{VAT-reordered dissimilarity matrix for Spotify dataset. No clear diagonal structure observed.}
    \label{fig:spotify}
\end{figure}

Despite a high Hopkins score (0.87), the VAT image and dimensionality reduction (PCA, t-SNE) show no clear clustering. This highlights VAT's advantage in visually invalidating misleading statistical indicators, especially in high-dimensional noisy datasets.

\subsubsection{Blobs Dataset}

\begin{figure}[H]
    \centering
    \includegraphics[width=0.95\linewidth]{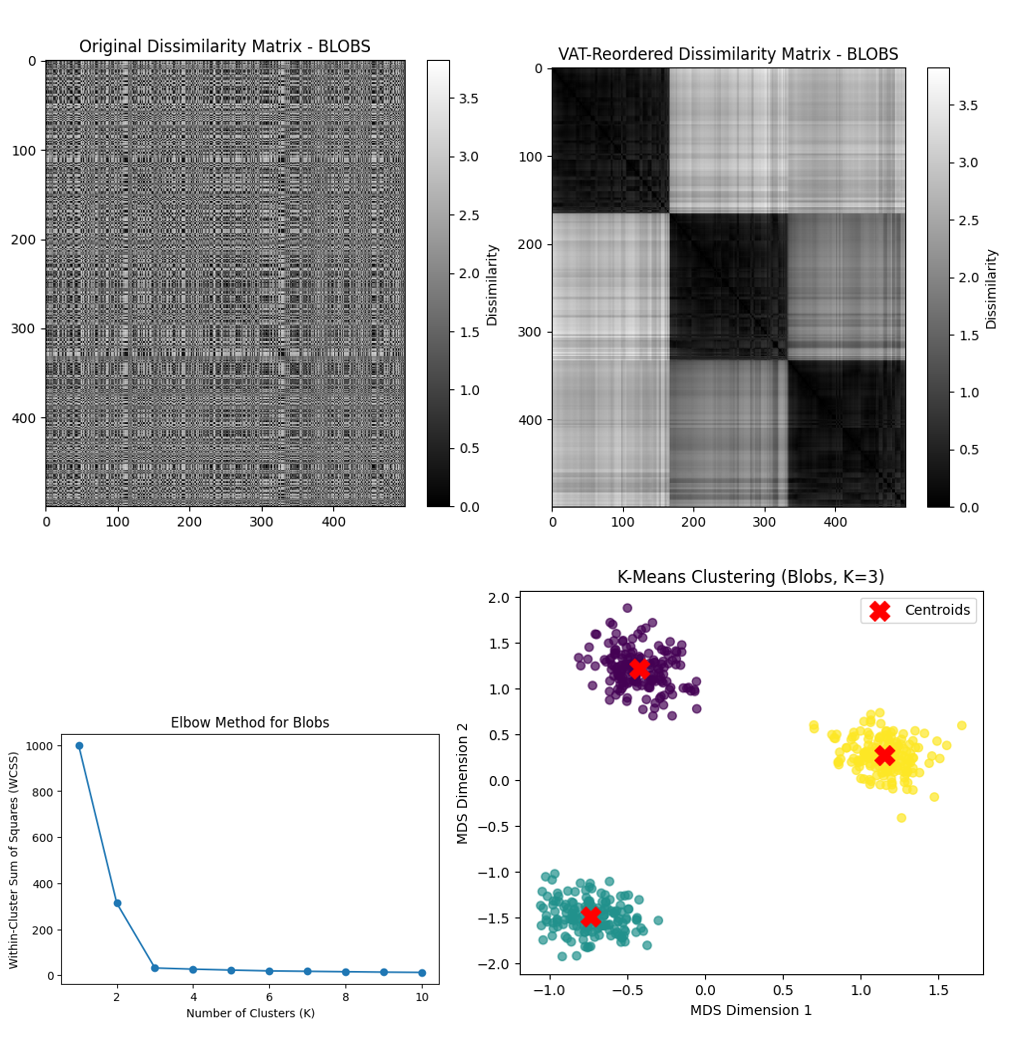}
    \caption{VAT image for the Blobs dataset. Strong diagonal blocks reflect well-separated Gaussian clusters.}
    \label{fig:blobs}
\end{figure}

Blobs is a synthetically generated dataset of spherical clusters. VAT, K-Means, and DBSCAN all align strongly. The Hopkins score of 0.93 further confirms high clusterability.

\subsubsection{Other Noteworthy Cases}

\textbf{Moons:} Non-linear crescents. VAT shows faint structure. K-Means fails; DBSCAN captures it perfectly. Hopkins: 0.89.  
\textbf{Circles:} Concentric ring challenge. VAT reveals weak structure. K-Means fails; DBSCAN succeeds. Hopkins: 0.73.  The Hopkins score of 0.73 for Circles is slightly below the 0.75 threshold, suggesting weak or borderline cluster structure, which aligns with VAT's indistinct diagonal blocks.
\textbf{GMM:} Overlapping Gaussians. VAT shows blurred diagonal. K-Means reasonably fits; DBSCAN is inconsistent. Hopkins: 0.94.

\section{Limitations and Future Work}

\subsection{Limitations}

Despite the significant speedups achieved through our Numba and Cython optimizations, several inherent limitations of the VAT algorithm remain unaddressed:

\begin{itemize}
    \item \textbf{Quadratic Memory Complexity:} VAT requires storage of the full pairwise dissimilarity matrix $R \in \mathbb{R}^{n \times n}$, resulting in $O(n^2)$ memory usage. This becomes a bottleneck for datasets with $n > 10^4$, especially on memory-constrained systems.

    \item \textbf{Sensitivity to Distance Metric:} The interpretability of the VAT image is closely tied to the choice of distance function. Our implementation assumes Euclidean distance, which may not capture relationships effectively in high-dimensional, sparse, or categorical data.

    \item \textbf{Computational Scalability:} Even with optimization, VAT remains an $O(n^2d)$ time complexity algorithm. This restricts real-time usage on large-scale datasets unless approximate or parallelizable variants are employed.

    \item \textbf{Limited Interpretability in Ambiguous Cases:} For datasets with weak or overlapping clusters, VAT images can be visually ambiguous, potentially leading to subjective interpretation errors.
\end{itemize}

\subsection{Future Work}

To address these limitations and further expand the applicability of VAT, we propose several avenues for future development:

\begin{itemize}
    \item \textbf{GPU-Accelerated Distance Computation:} Incorporating CUDA-enabled libraries such as RAPIDS cuML or PyTorch can accelerate the distance matrix computation, leveraging GPU parallelism for $O(1)$-time distance calculations per thread.

    \item \textbf{Approximate VAT via Sampling:} Inspired by sVAT~\cite{wu2007svat}, a subsampling-based strategy can significantly reduce time and memory requirements while preserving global structure. Techniques such as stratified or k-centroid sampling could be explored.

    \item \textbf{Dynamic or Learnable Distance Metrics:} Embedding distance metric learning (e.g., Mahalanobis, Siamese networks) within VAT could allow the algorithm to adaptively reflect data semantics, improving its cluster-revealing capacity across domains.

    \item \textbf{Pipeline Integration:} Developing a fully automated VAT+Clustering system where cluster tendency analysis directly informs the choice of clustering algorithm (e.g., selecting between K-Means and DBSCAN) could enhance unsupervised workflows.

    \item \textbf{Streaming VAT for Online Data:} Investigating incremental or streaming variants of VAT would allow it to handle continuous data flows, enabling real-time cluster tendency monitoring.
\end{itemize}

\section{Conclusion}

This study presents a high-performance implementation of the Visual Assessment of Cluster Tendency (VAT) algorithm using Python, Numba, and Cython. Our contributions include both algorithmic profiling and systematic optimization of the original VAT procedure, resulting in the following key outcomes:

\begin{itemize}
    \item The standard Python VAT implementation exhibits high computational cost and poor scalability on datasets beyond a few thousand points.
    \item A Numba-based JIT compilation approach yields consistent 25–35× speedups with minimal code refactoring.
    \item A Cython-based static compilation strategy achieves up to 50× acceleration by introducing explicit memory control and C-level data structures.
    \item Despite acceleration, the qualitative VAT outputs remain identical, preserving interpretability and diagnostic value.
    \item Visual and quantitative validations (e.g., Hopkins score, K-Means/DBSCAN comparisons) confirm the reliability of VAT for cluster tendency assessment.
\end{itemize}

While our work successfully mitigates runtime constraints, the $O(n^2)$ time and space complexity of VAT persists. Future directions include GPU-based parallelization, approximation via sampling, and the use of learnable or adaptive distance metrics to extend VAT's scalability and robustness to more challenging datasets.

To foster reproducibility and adoption, the optimized implementations are released as an open-source Python package, readily integrable into modern machine learning pipelines.

\subsection{Broader Impact}

Efficient cluster tendency analysis is critical for responsible unsupervised learning, yet often omitted due to computational overhead. Our accelerated VAT implementations enable the inclusion of this step in time-sensitive or large-scale domains such as:

\begin{itemize}
    \item \textbf{Healthcare and Genomics:} Rapid pattern recognition in gene expression or clinical cohorts~\cite{zhang2022bio}.
    \item \textbf{Finance and Anomaly Detection:} Real-time validation of customer segmentation and fraud detection pipelines~\cite{li2021finance}.
    \item \textbf{Recommendation Systems:} Dynamic user-group analysis in streaming environments.
\end{itemize}

By reducing latency while maintaining interpretability, our work promotes the deployment of VAT in high-throughput and high-stakes applications. The public availability of our package lowers barriers for research and industry adoption, contributing to transparent and verifiable clustering pipelines. In the broader context of AI, such tools are essential to ensuring that unsupervised models remain explainable, trustworthy, and aligned with practical constraints.

\subsection*{Code and Data Availability}
Our optimized VAT implementations are publicly available at: \url{https://github.com/Ashx098/VAT-Optimized}.  
All datasets used in this study (Iris, Spotify, Circles) are sourced from scikit-learn or open public repositories.







\end{document}